\documentclass{article} 
\usepackage{iclr2026_conference,times}

\usepackage{amsmath,amsfonts,bm}

\def\eqref#1{equation~\ref{#1}}

\def\1{\bm{1}}

\DeclareMathAlphabet{\mathsfit}{\encodingdefault}{\sfdefault}{m}{sl}
\SetMathAlphabet{\mathsfit}{bold}{\encodingdefault}{\sfdefault}{bx}{n}

\usepackage{hyperref}
\usepackage{url}
\usepackage{booktabs}
\usepackage{graphicx}
\usepackage{pifont}
\usepackage{hhline}
\usepackage[table]{xcolor}

\usepackage[most]{tcolorbox} 
\usepackage{xcolor}          

\definecolor{deepgreen}{RGB}{0,100,0}      
\definecolor{softgreen}{RGB}{80,200,120}    

\definecolor{deepblue}{RGB}{0,51,153}       
\definecolor{skyblue}{RGB}{70,130,180}     

\definecolor{deepred}{RGB}{153,0,0}       
\definecolor{softred}{RGB}{220,20,60}      

\definecolor{deeporange}{RGB}{204,85,0}  
\definecolor{softorange}{RGB}{255,140,0}   

\definecolor{deeppurple}{RGB}{102,0,153} 
\definecolor{softpurple}{RGB}{186,85,211}   

\definecolor{darkgray}{RGB}{80,80,80}      
\definecolor{lightgray}{RGB}{200,200,200}   

\definecolor{lightbluegray}{RGB}{232,240,247}
\usepackage{graphicx}
\usepackage{subcaption}

\title{UniETP: Unifying Environments for Generalizable Embodied Task Planning}

\author{Peiran Xu\thanks{Equal contribution}, Jiaqi Zheng\footnotemark[1], Ziyou Wang, Yadong Mu\thanks{Corresponding author} \\
Peking University 
}

\iclrfinalcopy 
\begin{document}

\maketitle

\begin{abstract}
This paper focuses on the problem of Embodied Task Planning, where an agent is required to execute a sequence of atomic actions within an interactive environment to complete a user-specified task. Though a variety of simulators and datasets have previously been built for this task, these efforts are largely isolated, with each using its own observation format, action type, and task domain. This fragmentation complicates comprehensive model evaluation and hinders the scalability of training data. As an effort towards generalizable embodied planning, we propose UniETP, a unified interface integrating four commonly-used simulators (AI2-THOR, VirtualHome, Habitat, BEHAVIOR). UniETP is characterized by both standardization and diversity. 
On one hand, it formalizes all the simulators into a consistent observation and action space, and builds an evaluation system to support complicated task goal.
On the other hand, it enhances task diversity and complexity across dimensions like task logic, instance grounding, and instruction understanding, constructing a new dataset with varied levels of difficulty in an automatic manner. Extensive experiments on the proposed benchmark are conducted to evaluate the embodied planning capabilities of recent models and analyze the performance bottlenecks.
Codes and data will be available at \href{https://github.com/woyut/UniETP}{https://github.com/woyut/UniETP}.
\end{abstract}

\vspace{-8mm}
\begin{figure}[htbp]
    \centering
    \includegraphics[
      width=0.95\linewidth,
    ]{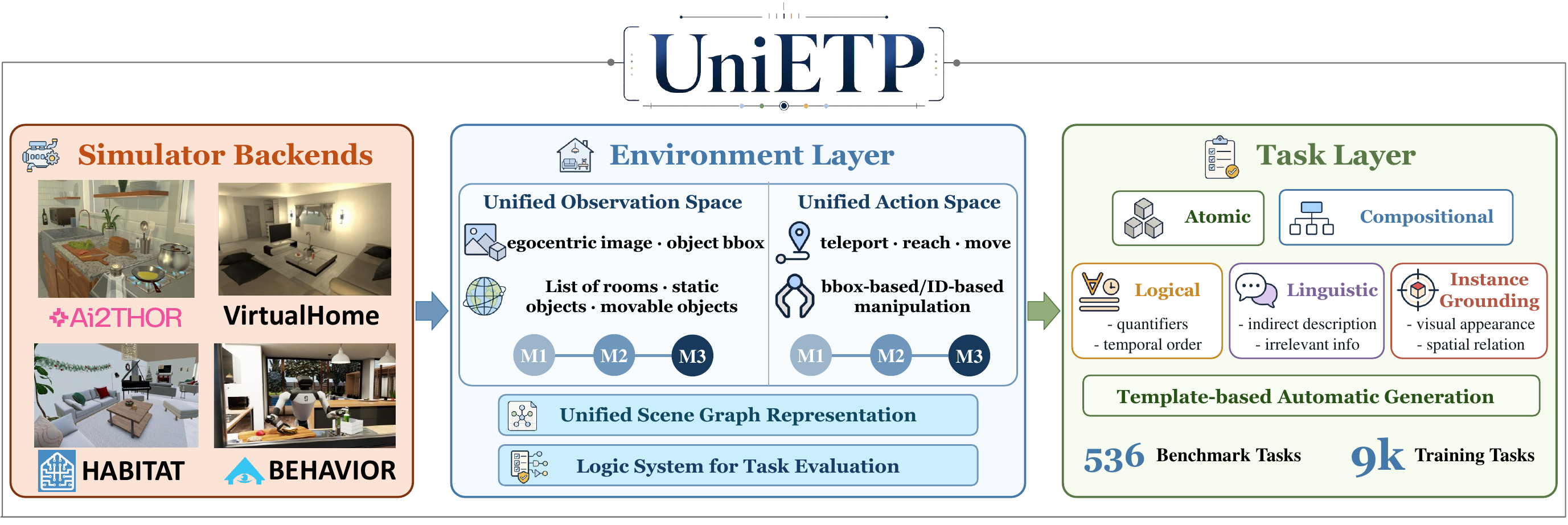}
    \vspace{-3mm}
    \caption{We propose UniETP, an embodied task planning benchmark consisting of an environment layer that unifies four simulators with standardized evaluation protocol, and a task layer that supports automatic generation of various task instances.}
    \vspace{-7mm}
\end{figure}

\section{Introduction}
\label{sec:intro}
\vspace{-3mm}

Recent advances in embodied intelligence have increasingly adopted a hierarchical architecture, in which high-level modules are responsible for task understanding and planning, while low-level controllers handle fine-grained motor execution~\citep{GR1.5,HiRobot,VLA-OS}. The problem of \textbf{Embodied Task Planning} (ETP) focuses on the former. It abstracts away detailed control and formulates decision-making at the level of atomic actions, requiring a model to interpret natural language task instructions and generate an appropriate sequence of actions to be executed in an interactive environment.
By bridging language understanding and action generation, ETP plays an important role in enabling embodied agents to accomplish complex, long-horizon tasks. As Vision–Language Models (VLMs) continue to advance, the study of ETP has attracted growing attention as a key component in the development of more capable and flexible robotic systems.

In recent years, a series of simulator environments~\citep{AI2-THOR,Habitat3.0,VirtualHome,Behavior} has been proposed or re-purposed for the problem of ETP, and many benchmarks and datasets have been built upon them~\citep{ALFWorld,VAB,EAI,EMBODIEDBENCH,PARTNR}. Despite the substantial progress enabled by these efforts, \textbf{isolation} across environments remains a critical challenge for embodied planning. Different simulators typically expose distinct environment information and implement actions in different ways. As a result, datasets built upon them often adopt different observation and action spaces, require planners to use different forms of inputs and outputs, and have different evaluation protocols or implementation details. Consequently, agents designed or tuned for one dataset are often difficult to transfer to another. Moreover, simulators are complementary in terms of scene scale, visual appearance, physical rules, object states, and supported interactions. Such isolation therefore also hinders the expansion of data scale and diversity, which are crucial for improving the generalization ability of embodied agents. 
Bridging the gap between simulators would allow models to be trained and evaluated on a more diverse set of tasks.

To address these challenges, we propose \textbf{UniETP}, a unified benchmark for developing and evaluating generalizable embodied agents. UniETP is organized into two levels: the environment layer and the task layer. At the \textbf{environment level}, UniETP unifies four widely used simulation environments through a Python-based environment interface that decouples agents from simulator-specific implementations. It further defines standardized, multi-granularity observation and action spaces, and adopts a unified scene graph representation as an environment abstraction to support unified evaluation logic. In this way, agents can operate under a consistent interface to complete tasks across different simulators. 

Building upon that, the \textbf{task level} supports the construction of diverse tasks with varying levels of difficulty. We define 100+ types of task templates, fully leveraging the rich object states and tool functionalities supported by the four simulators. Based on these templates, we develop an automatic task generation pipeline with the help of commonsense knowledge base and large language model (LLM) annotator. 
The resulting benchmark contains over 500 task instances, covering both simple atomic tasks and long-horizon compositional tasks. In designing these tasks, we pay particular attention to serveral capability dimensions, including instruction understanding, where instructions may involve complex references, syntactic structures, and distracting information; complex logic, where task goals may include temporal constraints, universal quantification, and other logical structures that go beyond the goal-state-based evaluation commonly used in prior work; instance-level grounding, where target objects are constrained to specific instances through states, spatial relations, or other conditions, in contrast to the category-level requirements adopted in many existing benchmarks.

In summary, the contributions of this paper are as follows:
\vspace{-2mm}
\begin{itemize}
    \item We unify four mainstream embodied simulators through a standardized environment interface, enabling the development of simulator-agnostic agents.
    \item We implement automatic generation of diverse tasks, allowing agents to be evaluated in a fine-grained manner across rich tasks, instructions, and scenes.
    \item We conduct extensive experiments on the proposed benchmark to examine the performance of recent VLMs in embodied planning.
\end{itemize}

\begin{table*}
  \caption{A comparison with previous benchmarks. Our UniETP unifies four popular embodied simulator (\textbf{A}I2-THOR, \textbf{V}irtualHome, \textbf{H}abitat, \textbf{B}EHAVIOR)). Combining their advantages, it supports visual modality (vis.) with rich object states (states). and a wide range of tasks (\#task types). It defines action space of multiple granularity (mul. act.) to support evaluation of different difficulty. We also propose an automatic task generation pipeline (auto. gen.), covering tasks with sophisticated instructions (instr.), complex goal logic (logic), instance-level grounding constraints (inst.). 
  *For some benchmarks that contain multiple subsets focusing on different domains, we only consider the subsets related to ETP.}
  \label{tab:bench}
  \vspace{-2mm}
  \centering
  \scalebox{0.88}{
  \begin{tabular}{l|ccccccccc}
    \toprule
Benchmark & sim. & vis. & \#task types & states & instr. & logic & inst. & mul. act. & auto. gen. \\
 \midrule
 ALFRED & A & \ding{51} & 7 & \ding{51} & \ding{55} & \ding{55} & \ding{55} & \ding{55} & \ding{55} \\
 ALFWorld & A & \ding{51} & 6 & \ding{51} & \ding{55} & \ding{55} & \ding{55} & \ding{55} & \ding{55}\\
 VisualAgentBench* & B & \ding{51} & 45 & \ding{51} &  \ding{55} & \ding{51} & \ding{55} & \ding{55} & \ding{55}  \\  
 LoTa-Bench & A,V & \ding{55} & 7+5 & \ding{51} & \ding{55} & \ding{55} & \ding{55} & \ding{55} & \ding{55}  \\
 ET-Plan-Bench & V & \ding{55} & 6 & \ding{55} & \ding{55} & \ding{51} & \ding{51} & \ding{55} & \ding{51} \\ 
  LEGENT & A & \ding{51} & 6 & \ding{55} & \ding{55} & \ding{55} & \ding{55} & \ding{55} & \ding{51} \\ 
 LangR & H & \ding{51} & 10 & \ding{55} & \ding{51} & \ding{51} & \ding{51} & \ding{55} & \ding{51} \\ 
 EAI & V,B & \ding{55} & 26+100 & \ding{51} & \ding{55} & \ding{51} & \ding{55} & \ding{55} & \ding{55} \\ 
 EmbodiedBench* & A,H & \ding{51} & 7+10 & \ding{51} & \ding{51} & \ding{51} & \ding{51} & \ding{55} & \ding{55} \\ 
 PARTNR & H & \ding{51} & - & \ding{55} & \ding{51} & \ding{51} & \ding{51} & \ding{55} & \ding{51}\\ 
 \midrule
 UniETP & A,V,H,B & \ding{51} & 138 & \ding{51} & \ding{51} & \ding{51} & \ding{51} & \ding{51} & \ding{51} \\
  \bottomrule
  \end{tabular}
  }
  \vspace{-6mm}
\end{table*}

\vspace{-3mm}
\section{Related Works}
\vspace{-1mm}
\subsection{Embodied Agent}
\vspace{-1mm}
Recent progress in foundation models has led to rapid advances in embodied intelligence, spanning across manipulation~\citep{CodePolicy,RT-2,PaLM-E,OpenVLA,pi0.5}, navigation~\citep{NaviLLM,NavGPT,NavFoM}, planning~\citep{LMZSP,LMIDM,LLMCK}, and reasoning~\citep{OpenEQA,SpatialVLM,Embodied-R}.
In particular, a growing body of work employs large models in an agentic manner, using them as high-level decision-making modules that serve as the ``brain'' of embodied agents and invoke navigation or control skills to accomplish complex tasks. Existing studies have explored designing environment and action representations~\citep{BUMBLE,ExploreVLM,CaPo,REVECA,ESCA}, incorporating memory modules and reflection mechanisms~\citep{InnerMonologue,LLM-Planner,Reflexion}, improving training strategies~\citep{Embodied-Reasoner,WAP,D2PO,MCTS-EP,SEEA-R1,RoboAgent}, and building embodied foundation models~\citep{GR1.5,RoboBrain2,Cosmos-Reason1}.

\vspace{-3mm}
\subsection{Simulators and Benchmarks for Embodied Task Planning}
\vspace{-2mm}
Environments are essential for the evaluation of embodied planner as it requires closed-loop interactions. The basis of such environments are interactive 3D simulators with scene and object assets. AI2-THOR~\citep{AI2-THOR} provides near-photorealistic indoor scenes with rich object properties and affordances, and its extensions~\citep{RoboTHOR,ProcTHOR} further support large-scale procedural scene generation and embodied manipulation. VirtualHome~\citep{VirtualHome,SEEAAS,WAH} focuses on simulating household activities through executable programs, offering a structured abstraction of human daily routines and object interactions. The Habitat series~\citep{Habitat,Habitat2.0,Habitat3.0} emphasizes scalable embodied AI research with efficient photorealistic simulation, configurable sensors, and support for diverse scene datasets. The BEHAVIOR series~\citep{Behavior,Behavior-1K} targets human-centered household activities with physical realism, including articulated objects, deformable bodies, and liquids. 
Other simulators~\citep{ThreeDWorld,GRUtopia,LabUtopia,CookBench,SayPlan} further broaden the landscape by focusing on larger-scale or domain-specific scenes.

On top of these simulators, embodied task planning benchmarks define evaluation protocols and task specifications that determine how agents interact with environments and what capabilities are tested. For example, ALFRED~\citep{ALFRED} introduces instruction following based on AI2-THOR, while ALFWorld~\citep{ALFWorld} further abstracts it with the TextWorld engine and defines a text-based agent interface. 
VisualAgentBench~\citep{VAB} considers various visual task domains for agents, including an embodied domain adapted from BEHAVIOR.
LoTa-Bench~\citep{LoTa-Bench} combines AI2-THOR and VirtualHome for evaluating language-oriented planners. ET-Plan-Bench~\citep{ET-PLAN-BENCH} utilizes VirtualHome for text-based tasks with spatial and temporal constraints. EAI~\citep{EAI} formalizes task planning into four stages and separately analyzes the performance of each.
LEGENT~\citep{LEGENT} and LangR~\citep{LLMGP} build task generation pipelines for communicable robots in AI2-THOR and mobile manipulation in Habitat.
EmbodiedBench~\citep{EMBODIEDBENCH} refines the tasks in LoTa-Bench and LangR, and combines them with low-level navigation and manipulation subsets. 
PARTNR~\citep{PARTNR} studies planning and reasoning in human-robot collaborative household tasks, with a large collection of tasks involving spatial, temporal, and heterogeneous constraints. Other efforts explore embodied agents in games~\citep{MineDojo,Octopus,UnrealZoo}, and safety-centric evaluations~\citep{SafeAgentBench,VestaBench,IS-Bench,HomeSafeBench}. 
A comparison among some typical previous benchmarks is presented in Table~\ref{tab:bench}.

\vspace{-3mm}
\section{The UniETP Framework}
\vspace{-1mm}
\subsection{Problem Formulation}
\vspace{-1mm}
The problem of Embodied Task Planning (ETP) can be formulated as a finite-horizon partially observable Markov decision process (POMDP) implemented by an underlying simulator, $\mathcal{E}=\langle \mathcal{S}, \mathcal{A}, \mathcal{O}, T, O, R, H \rangle$, where $\mathcal{S}$, $\mathcal{A}$, $\mathcal{O}$ are the spaces of states, actions, and observations, respectively. $T: \mathcal{S} \times \mathcal{A} \rightarrow \mathcal{S}$ denotes the environment transition dynamics. $O: \mathcal{S}\rightarrow\mathcal{O}$ is a mapping from states to observations. $R$ is a reward function. $H$ is the maximum number of action steps. 
Each task instance is defined by $\tau = \langle I, s_0, L,\mathcal{E} \rangle$, where $s_0 \in \mathcal{S}$ is the initial state, $L$ denotes the formalized task goal, and $I$ is an instruction describing the goal in natural language. The aim of ETP is to build a policy $\pi(a_t \mid h_t, I)$, typically implemented with a VLM that takes the instruction and interaction history $h_t=(o_0,a_1, o_1,...,o_{\text{t-1}})$ as input, and generates actions in the form of text. The agent receives a binary outcome reward at the end of the episode, indicating whether $L$ has been achieved.

\begin{figure}[htbp]
    \centering
    \includegraphics[
      width=0.97\linewidth,
    ]{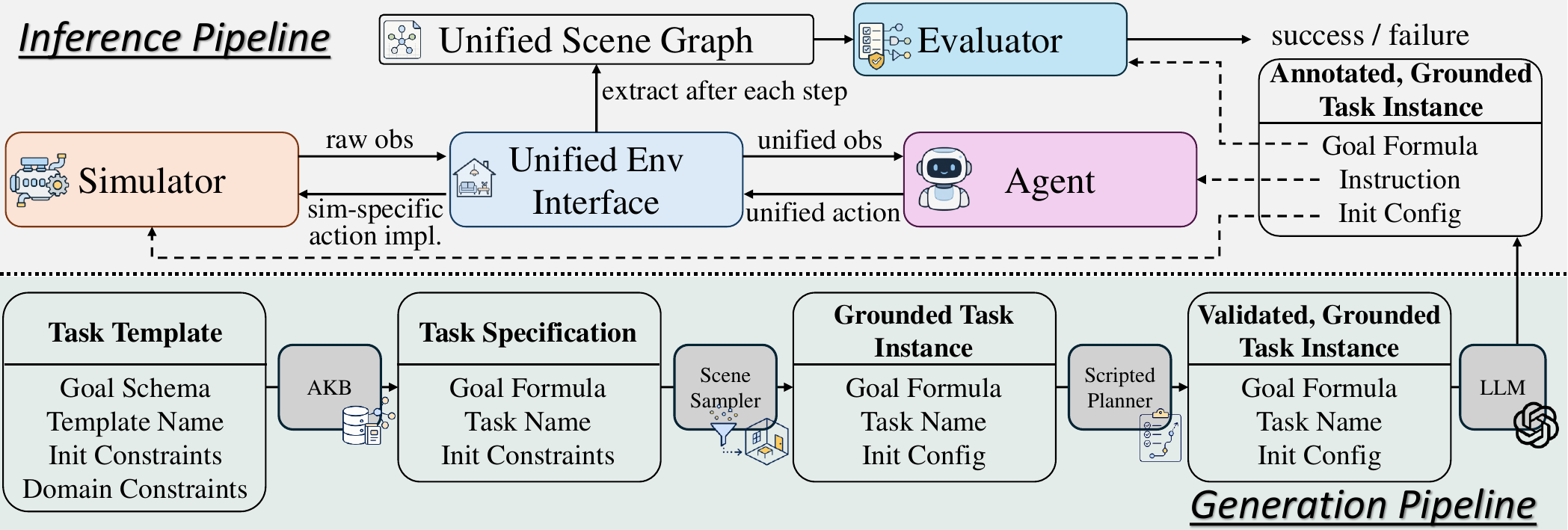}
    \vspace{-2mm}
    \caption{Upper: the ETP inference pipeline in UniETP's environment layer. Lower: the task generation process in UniETP's task layer. }\label{fig:pipeline}
    \vspace{-3mm}
\end{figure}

As discussed in Section~\ref{sec:intro}, for a set of simulators $\left\{\mathcal{E}^i\right\}_{i=1}^{n_E}$, each of them may have its own $\langle \mathcal{S}^i, \mathcal{A}^i, \mathcal{O}^i, T^i, O^i \rangle$, making the agent $\pi$ entangled with simulator-specific designs. The proposed UniETP addresses this problem by introducing a unified environment layer with a shared observation and action space $\langle \mathcal{O}^{\text{uni}} ,\mathcal{A}^{\text{uni}}\rangle$. The agent can thus receive simulator-agnostic observations $o^{\text{uni}}_t\in\mathcal{O}^{\text{uni}}$ and produce simulator-agnostic actions $a^{\text{uni}}_t\in\mathcal{A}^{\text{uni}}$, which are pre-processed from/post-processed to the simulator-compatible states/actions by the projectors in UniETP's environment layer, $\phi^i(s_t^i)=o^{\text{uni}}_t, \psi^i(a^{\text{uni}}_t)=a_t^i$.
The following subsections will explain the design of the unified environment layer and the task layer on top of it, and Figure~\ref{fig:pipeline} gives a visualization of the whole pipeline.

\vspace{-2mm}

\subsection{The Environment Layer}
\vspace{-1mm}
\subsubsection{Unified Interface}
\vspace{-1mm}
\underline{\textbf{Observation.}} UniETP defines a unified observation space that consists of three components: local view, global information, and feedback of the previous action, $o_t^{\text{uni}}=(l_t,g,f_t)$. The \textbf{local view} $l_t$ primarily consists of an egocentric RGB image captured by the agent's head-mounted camera. In addition, UniETP optionally provides a list of bounding boxes (bboxes) for objects visible in the current view, which allows us to disentangle the agent's basic visual perception and grounding skills from its task-oriented reasoning and planning capabilities.

The definition of \textbf{global information} varies across prior benchmarks. Some~\citep{ALFWorld} provide the agent with a list of objects in the scene at the beginning of an episode, while some~\citep{EMBODIEDBENCH} implicitly expose such information through an action space that includes ``\texttt{go to <sth>}" for all possible objects. We note that the content of global information has a significant impact on task difficulty in complex, cluttered environments. Access to more scene-level prior knowledge can enable the agent to complete tasks in a direct manner, while less information requires more deliberate exploration and environment understanding. To provide a clear and standardized protocol, UniETP organizes global scene information into three levels: rooms, static objects (\textit{e.g.}, furniture), and movable objects. By allowing $g$ to include some or all of these levels, UniETP controls the amount of scene priors available to the agent. This design enables systematic adjustment of task difficulty and facilitates analysis on the agent's exploration behavior.

The \textbf{feedback} $f_t$ represents the result of the previous action $a_{t-1}$. It is basically a binary signal indicating whether $a_{t-1}$ is successfully performed. For some commonly identifiable failure types (\textit{e.g.}, invalid action argument, object out of reach), we also include a short explanation in the feedback message to help the agent know why it fails. 

\underline{\textbf{Action.}} We represent each action instance with an action type plus zero to more objects/rooms as arguments. This parameterized formulation avoids enumerating all possible action instances~\citep{LLMGP,EMBODIEDBENCH}, thereby mitigating the combinatorial growth of the action space for scenes with a large number of objects.
To construct $\mathcal{A}^{\text{uni}}$, we examine the object states and action interfaces supported by different simulators, align actions with equivalent semantics, and merge them into a shared set of 30+ action categories. They can be broadly divided into navigation actions and manipulation actions. For \textbf{navigation}, UniETP supports three levels of action granularity. \textit{Movement} consists of basic locomotion actions, such as moving and rotating. \textit{Reaching} consists of turning and getting close to an object currently visible in the view. \textit{Teleportation} consists of turning and getting close to any specified object or room in the scene. These different levels of navigation abstraction affect the difficulty of object search and scene exploration.

For \textbf{manipulation}, UniETP distinguishes between \textit{atomic} actions and \textit{composite} actions. The former are non-decomposable primitive operations, whereas the latter can be realized by multiple atomic steps. For example, the action \texttt{heat with} (\texttt{potato}, \texttt{microwave}) can be decomposed into opening the microwave, placing the potato inside, closing and turning on the microwave, and so on. Introducing composite actions allows agents to bypass planning over certain common sub-procedures, which is another mechanism for controlling task difficulty.

Finally, we add \texttt{stop} to the action space, asking the agent to determine whether the task is completed instead of waiting for an oracle completion signal. The episode will also be terminated when the number of steps reaches $H$ or when the agent has tried $H'$ failed actions.

\underline{\textbf{Scene graph and evaluation logic.}} Apart from the unified observations and actions, UniETP defines a unified scene graph (USG) as a simulator-agnostic environment representation. In the graph, each node represents an object or a room, and stores its ID, category, intrinsic properties (\textit{e.g.}, \texttt{openable}), current states (\textit{e.g.}, \texttt{closed}), and spatial location. The agent is represented as a special node in the graph. Each edge encodes a spatial relation between two nodes, such as \texttt{on}, \texttt{inside}, \texttt{grasping}. The USG is constructed from the privileged information provided by the simulator and is updated after each executed action. It does not interact with the agent, but it serves as the basis for task evaluation and data generation.

For evaluation, we define a logic system over the USG traces. The logic combines first-order predicates over individual USGs (unary predicates checking whether a node has a certain state, and binary ones checking whether a certain edge exists between two nodes), Boolean connectives (and, or, not), first-order quantifiers (for all, exists), temporal operators over USG sequences (at end, until, before), and counting aggregates with numeric comparisons. Each task instance is associated with a goal formula $L$ built by these components. 

\vspace{-2mm}
\subsubsection{Evaluation Protocol}\label{sec:eval_protocal}
\vspace{-1mm}

\begin{table}[]
    \centering
    \caption{A comparison of the observation and action format among different evaluation modes.}\label{tab:modes}
    \vspace{-3mm}
    \scalebox{0.8}{
    \begin{tabular}{c|cc|ccc|ccc|cc}
    \toprule
        Eval. & \multicolumn{2}{|c|}{Local Obs} &  \multicolumn{3}{c|}{Global Obs} &  \multicolumn{3}{|c|}{Nav Action} &  \multicolumn{2}{c}{Manip Action} \\
        Mode & Img. & Bbox & Rooms & Stat. Obj. & Mova. Obj. & Teleport & Reach & Move & Atom. & Comp. \\
    \midrule
        M1 & \ding{51} & \ding{51} & \ding{51} & \ding{51} & \ding{51} & by ID & by ID & \ding{51} & by ID & by ID \\
        M2 & \ding{51} & \ding{51} & \ding{51} & \ding{51} & \ding{55} & by ID & by ID & \ding{51} & by ID & \ding{55}\\
        M3 & \ding{51} & \ding{55} & \ding{51} & \ding{55} & \ding{55} & by ID {\scriptsize (room only)} & by bbox & \ding{51} & by bbox & \ding{55} \\
    \bottomrule
    \end{tabular}
    }
\vspace{-7mm}
\end{table}

Based on the standardized multi-granularity observation and action spaces, UniETP defines three evaluation modes with varying difficulty, denoted as M1, M2, and M3, as illustrated in Table~\ref{tab:modes}. 
From the observation perspective, the global information $g^{\mathrm{M1}}$ contains a list of all rooms and object instances in the scene. $g^{\mathrm{M2}}$ contains rooms and static objects, while $g^{\mathrm{M3}}$ has only rooms. The local view $l_t^{\mathrm{M1}}$ and $l_t^{\mathrm{M2}}$ include a list of bboxes for visible objects, where each box is associated with an object ID and the pixel coordinates in the observation image. In contrast, $l_t^{\mathrm{M3}}$ has only the image.

From the action perspective, M1 allows the use of all action categories, including teleportation/reaching/movement-based navigation, and atomic/composite manipulation. The action arguments are specified by object/room IDs. M2 removes composite manipulation actions, and restricts the target of teleportation to rooms and static objects, in accordance with the information available in $g^{\mathrm{M2}}$. Actions in M2 are still parameterized by object/room IDs. M3 further restricts teleportation to rooms only, consistent with $g^{\mathrm{M3}}$. Moreover, except for teleportation-style navigation, all object-targeted actions in M3 must specify their argument using bboxes. We compare the model-generated bbox with the visible objects by computing the intersection-over-union (IoU), and executes the action on the best matched object. Therefore, M3 advances from choosing the target objects from the list of bboxes to actively locating them.

From M1 to M3, the amount of scene prior gradually decreases, the set of accessible actions becomes more restricted, and the burden of perception, localization, and exploration increases. Consequently, both the overall difficulty and the expected number of action steps increase. Intuitively, M3 simulates planning in a largely unfamiliar environment: the agent only has access to the basic room layout, while object placements and spatial relations are initially unknown. It must actively explore the environment, perceive objects, and accumulate local observations to build a sufficient understanding of the scene for task completion. M2 represents an intermediate setting in which the agent has partial prior knowledge of the environment, for example through a pre-built scene graph or semantic map that records the locations of major, static entities. This setting is well aligned with household applications, where furniture are typically stationary and can be memorized, while smaller movable objects may be relocated by humans and thus require search. M1 further assumes an ideal setting that the locations of all objects are known and additionally provides composite actions as high-level skills, allowing the agent to focus more on reasoning over user intent and task logic.

\vspace{-2mm}
\subsection{The Task Layer}\label{sec:task} 
\vspace{-1mm}
\subsubsection{Task Generation Pipeline}\label{sec:task_gen}
\vspace{-1mm}

The task layer of UniETP provides a pipeline for generating task datasets. It starts from a manually designed set of task templates. Each task template consists of a task name, a goal schema, domain constraints, and initial state constraints. We define a task template as a parameterized task specification whose target object categories have not yet been instantiated, \textit{e.g.}, \texttt{open one <A>}, \texttt{place two <A> and one <B> inside <C>}, and \texttt{wash one <A> then heat it}. The goal schema is a parameterized goal formula in which the quantification domains are represented by placeholders (\textit{e.g.},  ``exists object \texttt{x} of category \texttt{A} that is open at the end" for \texttt{open one <A>}). The domain constraints specify which object categories are valid substitutions for each placeholder. For instance, \texttt{open one <A>} requires \texttt{A} to be openable, \texttt{place one <A> inside <B>} requires that \texttt{A} can fit in \texttt{B}. The initial state constraints specify the conditions that the initial scene must satisfy, expressed as a set of predicates over the initial USG. For example, \texttt{open one <A>} requires the USG to contain at least one \texttt{A} and all \texttt{A} are initially closed. 

The second step is to instantiate the placeholders in each task template with appropriate object categories. To this end, we construct a commonsense affordance knowledge base (AKB) for each simulator. Each AKB lists all the object categories supported by the simulator, the states and functionalities associated with each category, and the feasible spatial relations between any two categories.
These AKBs are built from the documentation and metadata provided by the corresponding simulators, supplemented by LLM queries when the information is incomplete. By combining AKB with the domain constraints defined in each task template, we can fill in the placeholders and convert the parameterized task template into a concrete task specifications, including the goal formula $L$.

The third step is to ground each task specification in an initial scene. For each task, we first select a simulator $\mathcal{E}^i$ and randomly initialize a scene. We then enforce the initial state constraints by adding missing task-relevant objects when necessary and setting their initial states and locations accordingly. Next, we employ a scripted expert policy $\pi^*$ to attempt task planning and retain only the tasks that can be successfully completed, thereby ensuring task feasibility. Finally, we use an LLM to generate natural-language instructions $I$ based on the task template name, the goal formula, and the object information in the instantiated scene. This process yields the final annotated, grounded task instance, $\tau=\langle I,s_0^i,L,\mathcal{E}^i \rangle$.

\begin{figure}[htbp]
    \centering
    \includegraphics[
      width=0.95\linewidth,
    ]{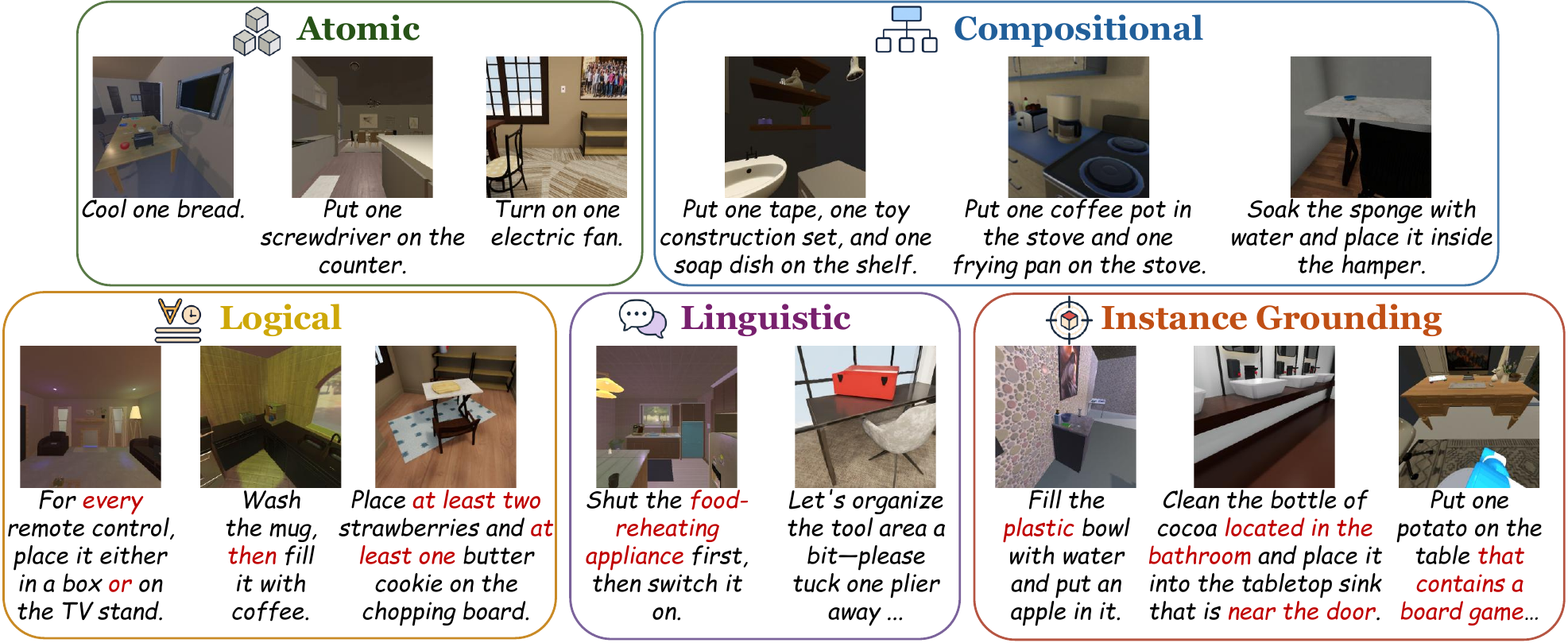}
    \vspace{-2mm}
    \caption{Example tasks on different splits.}\label{fig:dimension}
    \vspace{-3mm}
\end{figure}

\vspace{-2mm}
\subsubsection{Evaluation Dimensions}
\vspace{-1mm}
When constructing the benchmark tasks, we design several capability-oriented evaluation splits to examine different aspects of ETP. (1) \textbf{Atomic.} It contains the most basic task types with a single predicate as goal formula, evaluating the understanding of basic object states and spatial relations. (2) \textbf{Compositional.} It consists of the conjunctions of multiple atomic tasks, evaluating the ability to handle compositional requirements and perform long-horizon planning. (3) \textbf{Logical.} It contains tasks with explicit logical structures. Their goal formulas may involve universal quantification, temporal ordering, or counting aggregates and numerical comparisons. 
(4) \textbf{Linguistic.} It focuses on tasks with complex language instructions, 
which may refer to target objects through their functions, aliases, or contextual descriptions, requiring the agent to infer the intended object category. They may also introduce background information, distractors, or syntactic structures, requiring the agent to extract the actual objective from linguistically rich descriptions. (5) \textbf{Instance Grounding.} It imposes instance-level constraints on task-relevant objects. In contrast to the category-level requirements commonly used in prior datasets, (\textit{e.g.}, place a cup on a table), these tasks require the agent to select specific object instances based on visual attributes or spatial relations (\textit{e.g.}, place the \textit{dirty} cup on the \textit{wooden} table, place the cup \textit{in the coffee machine} to the table \textit{next to the chair}). Therefore, the agent must perform fine-grained visual recognition and localization, rather than relying only on object lists or bbox annotations. 

\begin{figure}[htbp]
\vspace{-1mm}
    \centering
    \includegraphics[
      width=0.95\linewidth,
    ]{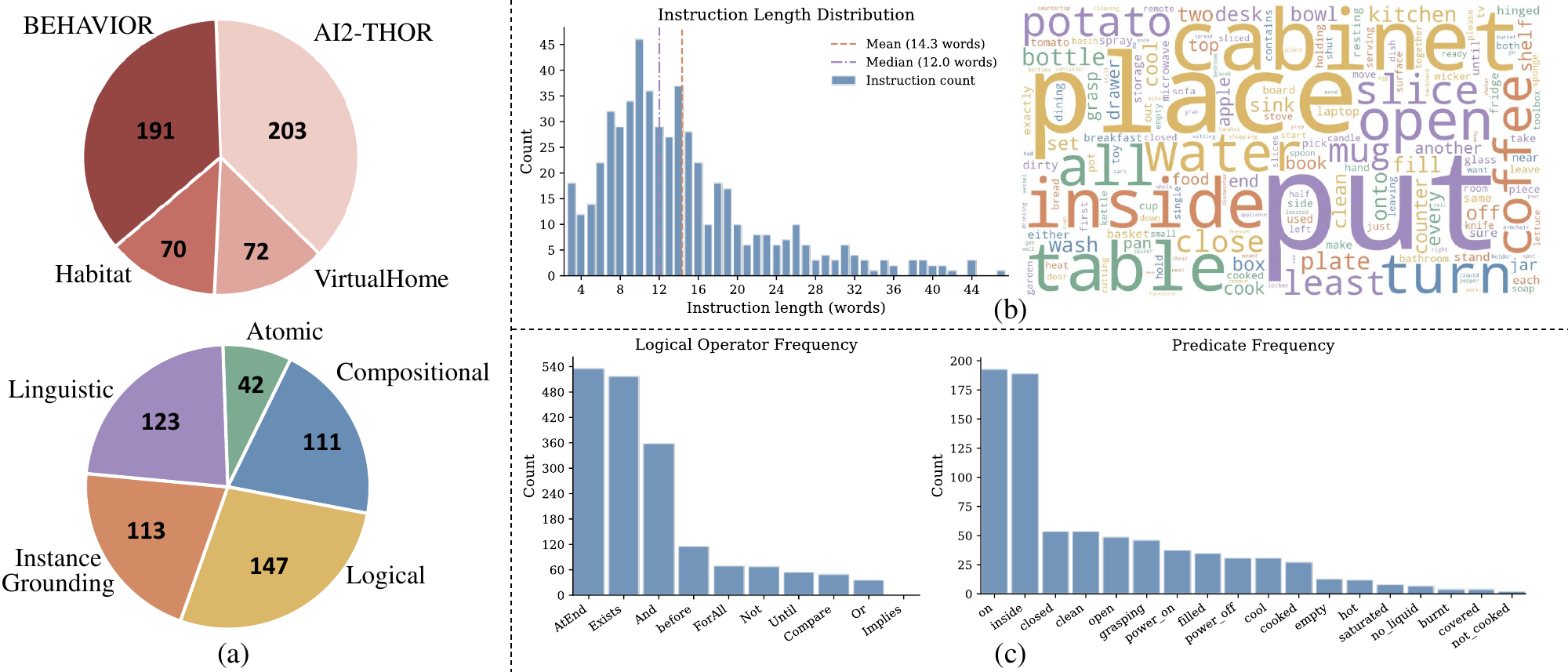}
    \vspace{-2mm}
    \caption{Key statistics of the UniETP benchmark. (a) Numbers of tasks across different simulators and splits. (b) Length distribution and word cloud of the instructions. The complex instructions in the Linguistic split account for the long tail of the distribution. (c) Frequencies of different logical operators and predicates, demonstrating the diversity of goal specifications and object states.}\label{fig:stat}
    \vspace{-3mm}
\end{figure}

\subsubsection{Benchmark Statistics}\label{sec:stat}
\vspace{-1mm}
The current version of UniETP combines AI2-THOR~\citep{AI2-THOR}, VirtualHome~\citep{VirtualHome}, Habitat~\citep{Habitat3.0}, and BEHAVIOR~\citep{Behavior-1K} into a unified interface. Fully leveraging the assets provided by the simulators, the benchmark covers 170 scene layouts. Each scene contains 7 rooms and 210 objects on average, and the objects come from the 125/181/389/1829 categories provided by each simulator. 
In terms of the action space, M1/M2/M3 contains 38/32/32 actions types, respectively, taking either object IDs or bboxes as arguments.

Through the pipeline described in Section~\ref{sec:task_gen}, we define 138 task templates and generate 536 task instances for the benchmark, covering a wide range of object states, spatial relations, and interaction patterns supported by the simulators. The feasibility is validated by the scripted expert policy and the language instructions are manually checked to ensure alignment with the task goal. The distributions of tasks, instructions, and goal formulas are presented in Figure~\ref{fig:stat}. We note that our generation pipeline has the potential to generate infinite grounded task instances. We build a larger training set of 9k tasks to support model fine-tuning, using a set of scenes that has no overlap with the benchmark. 

\vspace{-2mm}
\section{Experiments}
\vspace{-1mm}
\subsection{Setup}
\vspace{-1mm}
We benchmark a wide range of frontier models on UniETP, include leading proprietary models (GPT, Gemini, Claude), popular open models~\citep{Qwen3-VL,InternVL3.5,gemma4}, and embodied foundation models~\citep{ER1.6,RoboBrain,EmbodiedBrain,Cosmos-Reason1,Vlaser,RynnBrain,ACE-Brain,MiMo-Embodied}. We employ a ReAct-style~\citep{REACT} pipeline to evaluate each model in closed-loop interactions. In each calling, we include the description of the action space, the instruction, the global information, current local observation (image and bbox list for M1/M2, sole image for M3), and the action and feedback history into the input. The VLM is prompted to generate optional thinking and one or more actions in JSON format. 
The maximum steps $H$ and maximum failed steps $H'$ are set to 50/10 for M1 and M2, and 150/30 for M3. The task success rate (SR), evaluated with the goal formula, is used as the main metric.

\vspace{-2mm}
\subsection{Benchmark Results}
\vspace{-1mm}

\begin{table*}
  \caption{Performance comparison on UniETP benchmark under evaluation mode M1 and M2, including results across the atomic (At.), composite (Co.), logical (Lo.), instance grounding (In.), linguistic (Li.) splits and the average (Avg.) on the whole benchmark.}
  \vspace{-3mm}
  \label{tab:main}
  \centering
  \scalebox{0.92}{
  \begin{tabular}{l|ccccc>{\columncolor{lightbluegray}}c|ccccc>{\columncolor{lightbluegray}}c}
    \toprule
       & \multicolumn{6}{c|}{M1} & \multicolumn{6}{c}{M2}  \\
      & At. & Co. & Lo. & In. & Li. & Avg.  & At. & Co. & Lo. & In. & Li. & Avg.  \\
    \midrule
GPT-5.5 & 90 & 82 & 76 & 48 & 74 & 72 & 45 & 21 & 33 & 20 & 27 & 27 \\
Gemini-er-1.6 & 93 & \textbf{86} & \textbf{87} & \textbf{50} & 73 & \textbf{76} & 45 & 23 & 33 & 30 & 24 & 29 \\
Gemini-3.1-pro & \textbf{98} & 79 & 80 & 46 & 74 & 73 & \textbf{55} & \textbf{31} & 33 & \textbf{35} & \textbf{34} & \textbf{35} \\
Claude-sonnet-4.6 & 90 & 81 & 78 & 45 & 76 & 72 & 45 & 26 & 27 & 25 & 26 & 27 \\
Claude-opus-4.7 & \textbf{98} & 80 & 80 & \textbf{50} & \textbf{80} & 75 & 52 & \textbf{31} & \textbf{36} & 29 & 30 & 33 \\
\midrule
Qwen3.5-4B & 67 & 46 & 41 & 24 & 43 & 41 & 29 & 5 & 10 & 11 & 13 & 11 \\
Qwen3.5-9B & 76 & 68 & 62 & 37 & 60 & 59 & 33 & 10 & 14 & 12 & 15 & 14 \\
Qwen3.6-27B & 86 & 72 & 68 & 42 & 72 & 66 & 45 & 17 & 19 & 19 & 22 & 21 \\
InternVL3.5-8B & 43 & 21 & 19 & 13 & 21 & 21 & 19 & 3 & 9 & 4 & 11 & 8 \\
InternVL3.5-38B & 62 & 36 & 38 & 15 & 37 & 34 & 21 & 7 & 8 & 9 & 11 & 10 \\
Gemma4-E4B-it & 55 & 32 & 23 & 19 & 24 & 27 & 21 & 5 & 10 & 9 & 10 & 10 \\
Gemma4-31B-it & 86 & 67 & 72 & 40 & 68 & 64 & 40 & 13 & 22 & 19 & 23 & 21 \\
\midrule
RoboBrain2.5-8B & 36 & 19 & 22 & 12 & 24 & 21 & 24 & 5 & 11 & 4 & 8 & 9 \\
EmbodiedBrain-7B & 36 & 10 & 17 & 11 & 12 & 15 & 19 & 3 & 7 & 4 & 6 & 6 \\
Vlaser-8B & 45 & 9 & 7 & 6 & 11 & 11 & 19 & 3 & 3 & 4 & 4 & 5 \\
Cosmos-Reason2-8B & 36 & 17 & 15 & 9 & 20 & 17 & 21 & 5 & 7 & 5 & 6 & 7 \\
RynnBrain-8B & 43 & 17 & 16 & 12 & 18 & 18 & 17 & 3 & 5 & 5 & 5 & 5 \\
ACE-Brain-0-8B & 36 & 13 & 12 & 11 & 14 & 14 & 31 & 2 & 7 & 5 & 6 & 7 \\
MiMo-Embodied-7B & 57 & 15 & 14 & 10 & 14 & 17 & 24 & 3 & 9 & 7 & 8 & 8 \\
  
  \bottomrule
  \end{tabular}
  
  }
  \vspace{-3mm}
\end{table*}

\begin{table*}

\begin{minipage}{0.33\textwidth}
\centering
  \caption{Performance comparison under evaluation mode M3.}
  \vspace{-2mm}
  \label{tab:main_M3}
  \centering
  \scalebox{0.81}{
  \begin{tabular}{l|c}
    \toprule
      & At.   \\
    \midrule
GPT-5.5 & 26 \\
Gemini-3.1-pro & \textbf{40} \\
claude-opus-4.7 & 26 \\
  \bottomrule
  \end{tabular}
    }
\end{minipage}
\begin{minipage}{0.65\textwidth}
\centering
\caption{Performance comparison on different simulators}\label{tab:by_sim}
\vspace{-2mm}
\scalebox{0.81}{
  \begin{tabular}{l|ccccc|ccccc}
    \toprule
    & \multicolumn{5}{c|}{M1} & \multicolumn{5}{c}{M2} \\
      & A & V & H & B & Avg, & A & V & H & B & Avg.   \\
    \midrule
Gemini-3.1-pro & 64 & 82 & 70 & 79 & 73 & 46 & 64 & 27 & 15 & 35\\
Qwen3.6-27B & 56 & 69 & 69 & 73 & 66 & 28 & 51 & 11 & 7 & 21\\
RoboBrain2.5-8B & 15 & 40 & 19 & 20 & 21 & 7 & 28 & 1 & 5 & 9\\
  \bottomrule
  \end{tabular}
  }
\end{minipage}
\vspace{-2mm}
\end{table*}

The evaluation results under M1 and M2 are shown in Table~\ref{tab:main}. Overall, closed-source models achieve the strongest performance, with Gemini-er-1.6 obtaining the best average score in M1 and Gemini-3.1-pro in M2. Among open-source VLMs, increased model size leads to consistently better performance, and the gap between medium sized models (\textit{e.g.}, Qwen3.6-27B) and closed-source models is relatively modest. This saturation suggests that scaling model size may not be the only way to enhance planning capability. On the other hand, existing open-source embodied foundation models perform worse than general-purpose VLMs of comparable or even smaller sizes. The result implies that their embodied-specific fine-tuning may not transfer well to our closed-loop evaluation setting, which requires multi-turn interaction, complex instruction following, and long-horizon planning. 

Across task splits, Atomic tasks are the easiest as expected, and performance drops clearly on the Compositional split, showing that compound goals and extended horizon substantially increase the difficulty of planning. The Logical and Linguistic splits do not exhibit a consistent degradation relative to the Compositional split, meaning that understanding logical relations or complex linguistic descriptions may not be the dominant bottleneck for zero-shot VLM-based agent. Instance Grounding is the most challenging split for most models, with even the strongest models reaching only 50\% in M1. We notice that current agents struggle with fine-grained visual discrimination, and they primarily rely on categories when choosing target objects.

Comparing mode M1 and M2, we observe a substantial and consistent performance drop for all models. This confirms that M2 is considerably more challenging: the agent has to go through an explicit exploration process (teleportation to movable objects is banned), and plan with more primitive actions (composite actions are banned). For M3, we evaluate three closed-source models on the Atomic split only, as shown in Table~\ref{tab:main_M3}. We can observe an additional increase in difficulty and that even these basic tasks are challenging for the agents, since they must make more detailed navigation decisions and ground target objects by themselves. These results suggest that there remains substantial room for improving VLM-based embodied agents, through, for example, better scene memory, spatial modeling, object grounding, and reflection mechanisms.

\vspace{-2mm}
\subsection{Fine-Grained Analysis on Performance}
\begin{figure*}
  \centering
  \begin{minipage}{0.48\textwidth}
    \centering
    \includegraphics[
      width=\linewidth,
    ]{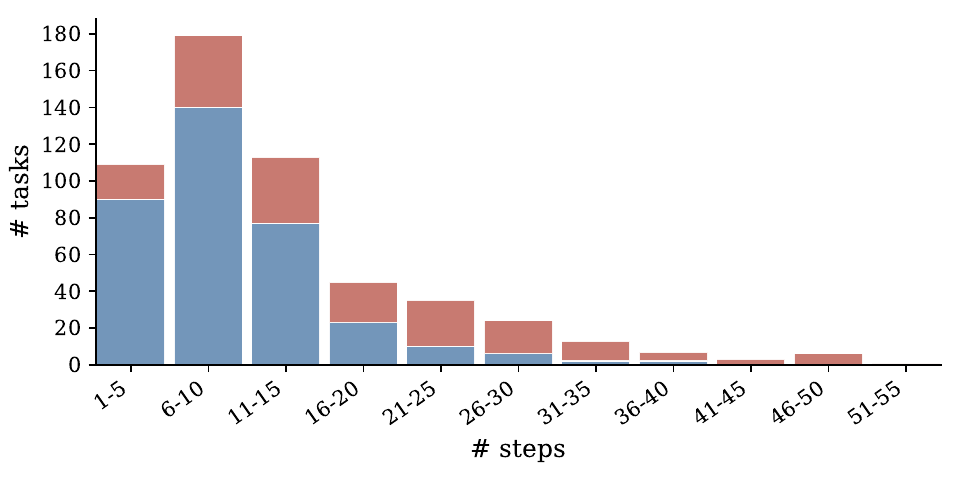}
  \end{minipage}%
  \hfill
  \begin{minipage}{0.48\textwidth}
    \centering
    \includegraphics[
      width=\linewidth,
    ]{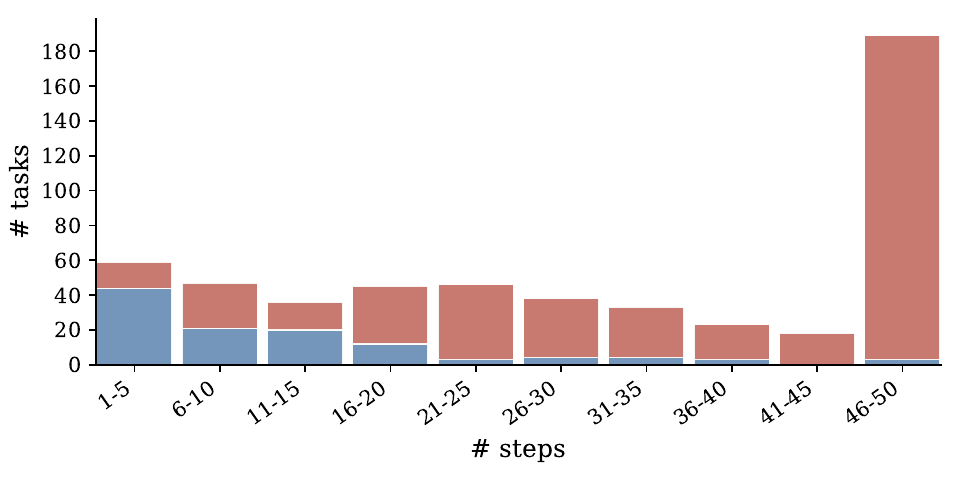}
  \end{minipage}
  \vspace{-3mm}
  \caption{The distribution of number of steps under mode M1 (left) and M2 (right).}
    \label{fig:step_M1M2}
    \vspace{-2mm}
\end{figure*}

\vspace{-1mm}
We further analyze model performance across different simulator backends, as shown in Table~\ref{tab:by_sim}. VirtualHome is the easiest domain in our benchmark, mainly because it contains fewer interactable objects and exposes a smaller set of executable action categories. Under M1, AI2-THOR presents the highest difficulty, as its tasks involve richer object states and more diverse tool affordances. However, under M2, performance on Habitat and BEHAVIOR drops substantially. This is largely due to their larger scene scales: agents must search for the small, movable target objects by traversing multiple rooms, making object discovery and exploration considerably more challenging. Figure~\ref{fig:step_M1M2} shows the distribution of episode lengths and the success ratio within each length range (using Qwen3.6-27B). The performance decreases steadily as the number of steps increases, indicating that models become less reliable when tasks require longer horizons and larger interaction context. In addition, due to the differences in the granularity of action space, M2 incurs a larger step cost than M1. Also, a lot of M2 episodes reach the maximum step limit, typically when the agent fails to find all the target objects.

\begin{table*}
  \caption{Ablation study on the effect of action success signal (Su.) and failure feedback (Fe.).} 
  \vspace{-1mm}
  \label{tab:abl_FSB}
  \centering
  \scalebox{0.82}{
  \begin{tabular}{cc|ccccc>{\columncolor{lightbluegray}}c|ccccc>{\columncolor{lightbluegray}}c}
    \toprule
    & & \multicolumn{6}{c|}{M1} & \multicolumn{6}{c}{M2}  \\
    Su. & Fe. & At. & Co. & Lo. & Li. & In. & Avg.  & At. & Co. & Lo. & Li. & In. & Avg.  \\
    \midrule

\ding{55} & \ding{55}  & 81 & 47 & 48 & 28 & 60 & 48 & 24 & 9 & 12 & 12 & 15 & 13  \\
\ding{51} & \ding{55}  & 83 & 75 & 69 & 43 & 71 & 66 & 40 & 16 & 21 & 19 & 20 & 21\\
\ding{51} & \ding{51}  & 86 & 72 & 68 & 42 & 72 & 66 & 45 & 17 & 19 & 19 & 22 & 21 \\
  \bottomrule
  \end{tabular}
  }
  \vspace{-3mm}
\end{table*}

\begin{table*}
  \caption{Performance of the SFT models. The regions colored in gray represent splits that require domain generalization.}
  \vspace{-2mm}
  \label{tab:train}
  \centering
  \scalebox{0.85}{
  \begin{tabular}{l|cc|cc|cc|cc|c}
    \toprule
 & \multicolumn{2}{c|}{A} & \multicolumn{2}{c|}{V} & \multicolumn{2}{c|}{H} & \multicolumn{2}{c|}{B} & Avg. \\
 & Seen & Unseen & Seen & Unseen & Seen & Unseen & Seen & Unseen & \\
    \midrule
Qwen3.5-4B & 23 & 31 & 58 & 53 & 23 & 41 & 55 & 58 & 41 \\
SFT (w.o. AI2-THOR) & \cellcolor{gray!20}44 & \cellcolor{gray!20}38 & 68 & 84 & 81 & 82 & 73 & 69 & 62 \\
SFT (w.o. unseen tasks)& 55 & \cellcolor{gray!20}49 & 68 & \cellcolor{gray!20}63 & 74 & \cellcolor{gray!20}82 & 69 & \cellcolor{gray!20}64 & 63 \\
SFT (full dataset) & 59 & 54 & 68 & 84 & 79 & 94 & 74 & 72 & \textbf{68} \\

  \bottomrule
  \end{tabular}
  }
  \vspace{-2mm}
\end{table*}

Finally, Table~\ref{tab:abl_FSB} presents an ablation study on action feedback, using Qwen3.6-27B. Under both M1 and M2, removing textual failure feedback has little impact on the overall performance. We observe in experiments that strong models are often able to infer plausible causes of failure, whereas weaker models have limited ability to recover once a failure occurs. As a result, textual explanation is not always effectively utilized by the agent. In contrast, removing the binary action success indicator leads to a clear performance drop. This suggests that current agents still have difficulty inferring whether an action has been successfully executed solely from visual observations, highlighting the importance of explicit post-verification mechanisms in embodied agentic systems.

\vspace{-2mm}
\subsection{Analysis on Training and Generalization}
\vspace{-1mm}

In this subsection, we present a preliminary attempt to develop generalizable embodied agents using UniETP. Based on the training set constructed as in Section~\ref{sec:task}, we perform supervised fine-tuning (SFT) on Qwen3.5-4B. As shown in Table~\ref{tab:train}, the fine-tuned model achieves a significant performance improvement and even outperforms models with more parameters (Qwen3.6-27B, Gemma4-31B-it), demonstrating the reliability of our data generation and supervision construction pipeline.

Furthermore, we exclude all episodes from a specific simulator (AI2-THOR), or from a subset of task templates (about 1/3 of the full template set), in the training set. At test time, the resulting models still show clear improvements over the original pretrained model in the unseen simulator environments and the unseen task types. These results suggest that bridging simulator boundaries and unifying the observation and action interface can contribute meaningfully to the training of agents with stronger generalization capability.

\vspace{-2mm}
\section{Conclusion}
\vspace{-1mm}
In this work we propose UniETP, a unified benchmark for evaluating embodied planning agent across heterogeneous simulation environments. UniETP transforms four commonly-used simulators into a standardized observation and action space, and defines an evaluation protocol with adjustable difficulty. Furthermore, it introduces an automated task generation pipeline to construct a benchmark covering diverse capability dimensions, together with a large-scale training set. Comprehensive experiments across a wide range of models reveal insights into the performance and limitations of VLM-based agents in embodied scenarios.

\clearpage

\bibliography{iclr2026_conference}
\bibliographystyle{iclr2026_conference}

\end{document}